\documentclass[manuscript]{acmart}
\AtBeginDocument{%
  \providecommand\BibTeX{{%
    \normalfont B\kern-0.5em{\scshape i\kern-0.25em b}\kern-0.8em\TeX}}}
\copyrightyear{2020}
\acmYear{2020}
\setcopyright{acmcopyright}\acmConference[RecSys '20]{Fourteenth ACM Conference on Recommender Systems}{September 22--26, 2020}{Virtual Event, Brazil}
\acmBooktitle{Fourteenth ACM Conference on Recommender Systems (RecSys '20), September 22--26, 2020, Virtual Event, Brazil}
\acmPrice{15.00}
\acmDOI{10.1145/3383313.3412216}
\acmISBN{978-1-4503-7583-2/20/09}

\usepackage{multirow}
\usepackage{graphicx}
\usepackage{stfloats}

\begin{document}
\title[MEANTIME]{MEANTIME: Mixture of Attention Mechanisms with Multi-temporal Embeddings for Sequential Recommendation}



\author{Sung Min Cho}
\affiliation{
  \institution{Seoul National University}
}
\email{tjdals4565@gmail.com}

\author{Eunhyeok Park}
\affiliation{
  \institution{POSTECH}
}
\email{canusglow@gmail.com}

\author{Sungjoo Yoo}
\affiliation{
  \institution{Neural Processing Research Center}
}
\affiliation{
  \institution{Seoul National University}
}
\email{sungjoo.yoo@gmail.com}

\renewcommand{\shortauthors}{Sung Min Cho, Eunhyeok Park, and Sungjoo Yoo}

\begin{abstract}
Recently, self-attention based models have achieved state-of-the-art performance in sequential recommendation task. Following the custom from language processing, most of these models rely on a simple positional embedding to exploit the sequential nature of the user's history. However, there are some limitations regarding the current approaches. First, sequential recommendation is different from language processing in that timestamp information is available. Previous models have not made good use of it to extract additional contextual information. Second, using a simple embedding scheme can lead to information bottleneck since the same embedding has to represent all possible contextual biases. Third, since previous models use the same positional embedding in each attention head, they can wastefully learn overlapping patterns. To address these limitations, we propose MEANTIME (MixturE of AtteNTIon mechanisms with Multi-temporal Embeddings) which employs multiple types of temporal embeddings designed to capture various patterns from the user's behavior sequence, and an attention structure that fully leverages such diversity. Experiments on real-world data show that our proposed method outperforms current state-of-the-art sequential recommendation methods, and we provide an extensive ablation study to analyze how the model gains from the diverse positional information.
\end{abstract}


\maketitle

\section{Introduction}

Capturing users' preferences from their history is essential for making effective recommendations, because users' preferences and items' characteristics are both dynamic. Furthermore, users' preferences depend heavily on the context (e.g., one is likely to be interested in buying a keyboard after purchasing a desktop). Therefore, sequential recommendation aims to predict the next set of items that users are likely to prefer by exploiting their history.

Many algorithms have been proposed to better understand the sequential history of users~\cite{SASRec, BERT4Rec, hidasi2018recurrent, CASER, M3, rendle2010factorizing, DIN, TRANSREC}. Despite their excellent performance, most of them ignore the interactions' timestamp values. While recent works such as TiSASRec~\cite{TISASRec} successfully incorporated time information, their usage of time was also limited to a single embedding scheme. Since the timestamp values are rife with information, it would be beneficial to explore many forms of temporal embeddings that can fully extract diverse patterns, and model architectures that can fully leverage such diversity.


In this paper we propose \textbf{MEANTIME} (\textbf{M}ixtur\textbf{E} of \textbf{A}tte\textbf{NTI}on mechanisms with \textbf{M}ulti-temporal \textbf{E}mbeddings) which introduces multiple temporal embeddings to better encode both absolute and relative positions of a user-item interaction in a sequence. Moreover, we employ multiple self-attention heads that handle each positional embedding separately. This is profitable because each head can play a role of expert that focuses on a particular pattern of user behaviors. 
Extensive experiments show that \textbf{MEANTIME} outperforms state-of-the-art algorithms on real-world datasets. Our contributions are as follows:
\begin{itemize}
\item We propose to diversify the embedding schemes that are used to encode the positions of user-item interactions in a sequence. This is done by applying multiple kernels to positions/values of timestamps in a sequence to create unique embedding matrices.
\item We also propose a novel model architecture that operates multiple self-attention heads simultaneously, where each head specializes in extracting certain patterns from the users' behaviors by utilizing one of the positional embeddings we propose above.
\item We conducted thorough experiments to show the effectiveness of our method on real-world datasets. We also provide a comprehensive ablation study that helps understand the effect of various components in our model.
\end{itemize}


\section{Related Work}

\subsection{Temporal Recommendation}


Since temporal information hold contextual information, they can be very crucial for recommendation performance. TimeSVD++ and BPTF~\cite{TIMESVD++, BPTF} adopted time factor into the matrix factorization method, and TimeSVD++ was one of the main contributions for the winning of Netflix Grand Prize~\cite{NETFLIXPRIZE}.
Time-LSTM~\cite{TIMELSTM} equipped LSTMs with several forms of time gates to better model the time intervals in user's interaction sequence. Recently, CTA~\cite{CTA} used multiple parametrized kernel functions on temporal information to calibrate the self-attention mechanism.

\subsection{Sequential Recommendation}

Sequential recommendation aims to suggest relevant items based on the user's sequential history. Markov-chain based methods~\cite{SIMILARITYMC, MCAULEYMC2016, TRANSREC, he2016fusing} assume that users' behaviors are affected only by their last few behaviors. FPMC~\cite{rendle2010factorizing} merges Markov-chains with matrix factorization method for next-basket recommendation. Since the first suggestion by GRU4Rec~\cite{GRU4Rec}, many RNN-based methods~\cite{hidasi2018recurrent, RNN4Rec, HIERARCHICALGRU4REC, DREAM, wu2017recurrent, wu2017sequential, DINEVOLUTION} brought the success of RNN into item sequence understanding. To overcome the strong order constraint of RNN models, CNN-based methods~\cite{CASER, CNNGENERATIVE, HIERARCHICALCNN} were proposed. Some works adopted graph neural network (GNN) to understand user's session as a graph~\cite{SESSIONGNN}.


As for attention mechanisms, NARM~\cite{NARM} and STAMP~\cite{STAMP} incorporated vanilla attention mechanism with RNN. More recently, SASRec~\cite{SASRec} successfully employed self-attention mechanism, which was a huge success in NLP areas~\cite{TRANSFORMER, BERT, TRANSFORMER-XL, XLNET}. BERT4Rec~\cite{BERT4Rec} improved SASRec by adopting Transformer~\cite{TRANSFORMER} and Cloze-task based training method. TiSASRec~\cite{TISASRec} enhanced SASRec by merging timestamp information into self-attention operations. Our work seeks to further expand this trend by suggesting a novel model architecture that uses multiple types of temporal embeddings  and self-attention operations to better capture the diverse patterns in user's behaviors.


\section{Proposed Method}
\begin{figure*}[t]
  \centering
  \includegraphics[width=0.9\linewidth]{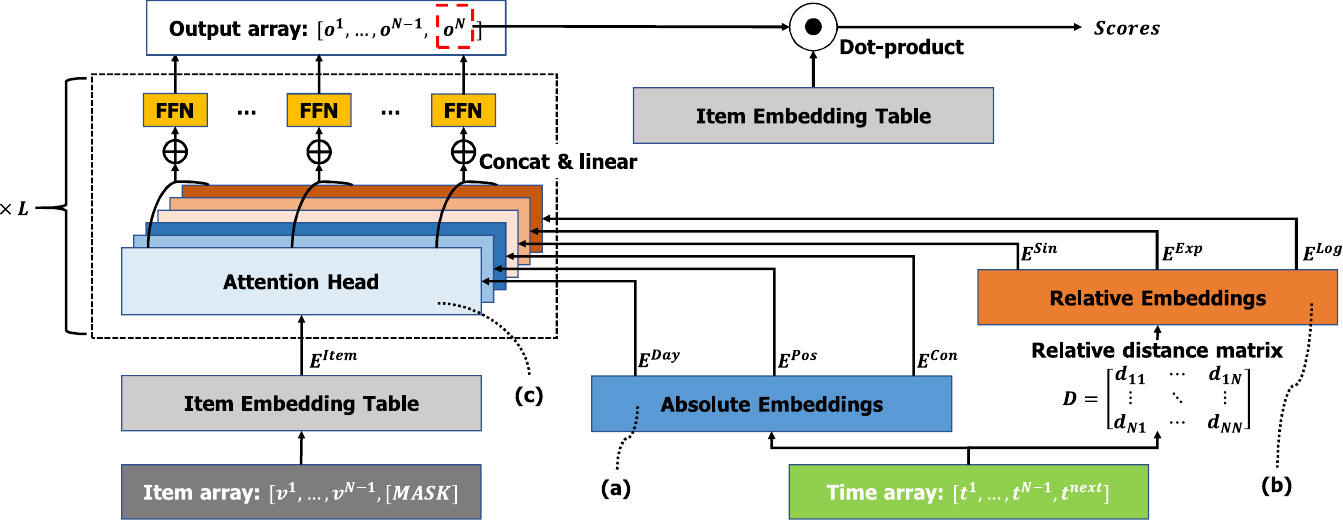}
  \caption{Model Architecture}
  \Description[The figure describes our model architecture.]{In the figure, item array is converted by item embedding table and fed to each attention head. Time array is converted by absolute/relative embeddings into 6 different embeddings. Each attention head receives item embedding and unique temporal embedding. Concat&linear and FFN operation lies above the heads. Heads and FFNs are repeated for L layers. At the end of the layer, output vector is dot-producted with item embedding table to compute the final item score distribution.}
  \label{fig:model}
\end{figure*}

\begin{figure*}[t]
  \centering
  \includegraphics[width=\linewidth]{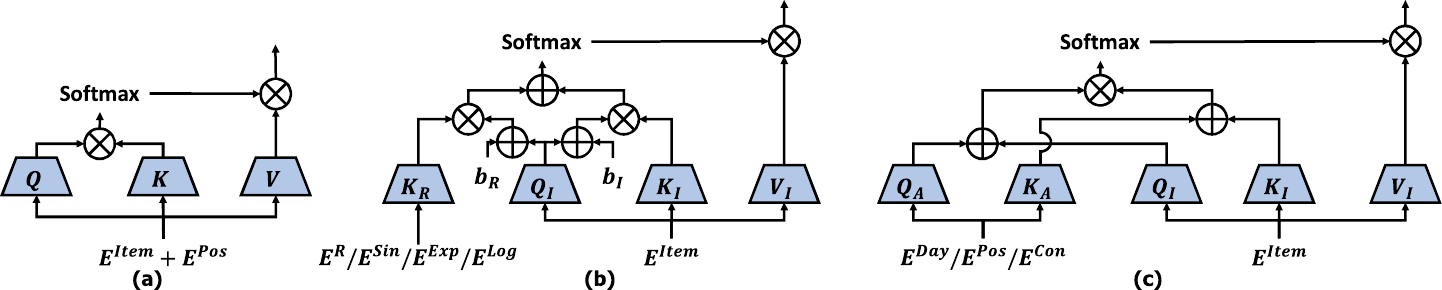}
  \caption{Brief overview of single attention head in multi-headed attention structure. (a) absolute self-attention in Transformer (b) relative self-attention in Transformer-XL and MEANTIME (c) absolute self-attention in MEANTIME.}
  \Description[The figure describes three different types of single attention-head]{Figure(a) describes a single attention head used in transformer. The sum of E^Item and E^Pos is fed to the head. Figure(b) describes a single attention head from Transformer-xl and our model. E^Pos is replaced by E^R and positional information is not used in value computation. Linear layers are also separated between content and position. Figure(c) describes a single attention head used for absolute attention in our model. It borrows the structure from (a), but separates the content and position as in (b).}
  \label{fig:attntypes}
\end{figure*}

Several self-attention based recommendation models have been proposed recently~\cite{SASRec, BERT4Rec}. However, the positional factors have been rather overlooked in these models for many reasons. First, they don't utilize timestamp values which hold important contextual information. While TiSASRec~\cite{TISASRec} successfully addressed this issue, they also used a simple embedding scheme for temporal values. This can possibly lead to an information bottleneck since the same embedding has to represent all possible positional biases. Furthermore, since all attention heads use the same positional information, they might wastefully learn the same patterns. As a way to mitigate these problems, we devise a novel model architecture that injects various temporal embeddings for each attention head as shown in Figure \ref{fig:model}.

\subsection{Problem Formulation}

Let $U$ be a set of users, and $V$ a set of items. For each user $u\in U$, we have a sequence of items $V^u=[v^u_1, ..., v^u_k, ..., v^u_{|V^u|}|v^u_k\in V]$ that the user previously interacted with in chronological order and the corresponding time sequence of the interaction $T^u=[t^u_1, ..., t^u_k, ..., t^u_{|V^u|}|t^u_k\in \mathbb{N}]$ that stores the absolute timestamp values.
Our goal is to predict the next item $v^u_{next}$ that the user $u$ is likely to interact with at the target timestamp $t^u_{next}$ based on the given history $(V^u, T^u)$. 

\sloppy Our model is based on a fixed-length model with length $N$ as in previous works~\cite{SASRec, BERT4Rec}. Given the history $(V_u, T_u)$ of user $u$, we feed $\mathbf{v} = [v^u_{|V^u|-N+2}, ..., v^u_{|V^u|}, \text{[MASK]}]$ and $\mathbf{t} = [t^u_{|V^u|-N+2}, ..., t^u_{|V^u|}, t^u_{next}]$ to the model. Then the model estimates $v_{next}$ as output. If the history is shorter than $N-1$, it is padded by a special token [PAD].

\subsection{Input Embedding}
\label{section:inputembedding}
The embedding layers convert $\mathbf{v}$ and $\mathbf{t}$ to hidden features that are provided to the attention module. In the case of item array $\mathbf{v}$, each element has an item index. This array is embedded as $E^{item} \in \mathbb{R}^ {N\times h}$ using the conventional look-up operation with a learnable item embedding table $M^I\in \mathbb{R}^{(|V|+1)\times h}$, where h is the hidden feature size. $M^I$ holds embeddings for $|V|$ items and [MASK] token.

In the case of $\mathbf{t}$, each element has a timestamp value.
In previous studies, these values were embedded in a single embedding scheme, if they were used at all. However, given that user's behaviors are mixed with various patterns, a single embedding might not suffice. Moreover, using different encoding functions (instead of plain learnable values) for each embedding is beneficial since we can train each of them to learn a unique pattern. With these in mind, we propose six methods to embed $\mathbf{t}$: three absolute methods (Figure \ref{fig:model}(a)), and three relative methods (Figure \ref{fig:model}(b)).

Absolute embeddings encode each position in the sequence as $E^{Day}, E^{Pos}\text{, or } E^{Con} \in \mathbb{R}^{N \times h}$. The first is $\textbf{Day}$-embedding which converts the day of each timestamp into an embedding vector to get $E^{Day}$. We keep a learnable embedding matrix $M^D\in R^{|D|\times h}$ where $|D|$ is the number of possible days in the span of the dataset. In our analysis, $\textbf{Day}$ is a reasonable choice of time length considering the trade-off between matrix size and the effectiveness. $\textbf{Pos}$-embedding is analogous to the learnable positional embedding used in~\cite{BERT, BERT4Rec}, where each position has a corresponding embedding in $M^P\in R^{N\times h}$. We convert each position to get $E^{Pos}$. Note that in this case, the embeddings depend only on the indices of time array and not on the timestamp values. \textbf{Con}-embedding is similar to \textbf{Pos}-embedding, except that all position vectors are shared ($M^C\in R^{1\times h}$) to remove positional bias. Hence, $E^{Con}$ is a stack of repeated vectors.

Relative embeddings encode the relationship between each interaction pair in the sequence as $E^{Sin}, E^{Exp}\text{, or } E^{Log} \in \mathbb{R}^{N \times N \times h}$ by utilizing the temporal difference information. First, we define a matrix of temporal differences $\mathbf{D} \in \mathbb{R}^{N \times N}$ whose element is defined as $d_{ab} = (\mathbf{t}_a-\mathbf{t}_b)/\tau$, where $\tau$ is an adjustable unit time difference. We introduce three encoding functions for \textbf{D}.
First, $\textbf{Sin}$ encoder converts the difference $d_{ab}$ to a hidden vector $\vec{\theta}_{ab} \in \mathbb{R}^{1 \times h}$, by the following equation:
\begin{equation}
    \vec{\theta}_{ab, 2c} = sin(\frac{d_{ab}}{freq^{\frac{2c}{h}}})\qquad\vec{\theta}_{ab, 2c+1} = cos(\frac{d_{ab}}{freq^{\frac{2c}{h}}})
\end{equation}
where $\vec{\theta}_{ab, c}$ is the $c^{th}$ value of the vector $\vec{\theta}_{ab}$ and $freq$ is an adjustable parameter.
Likewise, $\textbf{Exp}$ encoder and $\textbf{Log}$ encoder converts $d_{ab}$ to $\vec{e}_{ab}$ and $\vec{l}_{ab}$ respectively by applying the following equations:
\begin{equation}
  \vec{e}_{ab, c} = exp(\frac{-|d_{ab}|}{freq^{\frac{c}{h}}}) \qquad
  \vec{l}_{ab, c} = log(1 + \frac{|d_{ab}|}{freq^{\frac{c}{h}}})
\end{equation}
Stacking these vectors gives us $E^{Sin}, E^{Exp}$ and $E^{Log}$ respectively.

Each embedding offers a distinctive view on temporal data. For example, $\textbf{Sin}$ captures periodic occurrences. Larger time gap is either quickly decayed to zero in $\textbf{Exp}$ or manageably increased in $\textbf{Log}$. $\textbf{Day}$ can confine the attention scope within the same (or similar) day. $\textbf{Pos}$ can pick up the patterns learnt by the previous models. Finally, $\textbf{Con}$ removes positional bias altogether so that the attention can focus only on the relationship between items. These distinctive views enable our self-attention heads (Figure \ref{fig:model}(c)) to attend to different characteristics of the sequence.

\subsection{Self-Attention Structure}
\subsubsection{Attention Architecture}
Multi-head self-attention proposed in Transformer~\cite{TRANSFORMER} shows excellent performance in recommendation tasks~\cite{BERT4Rec}. MEANTIME is also based on this structure. Traditionally, self-attention based models~\cite{SASRec, BERT, BERT4Rec} employ a simple absolute positional encoding by feeding the sum of $E^{Item}$ and $E^{Pos}$ as input as shown in Figure \ref{fig:attntypes}(a). Advanced transformers~\cite{TRANSFORMER-XL, XLNET} enhanced this as shown in Figure \ref{fig:attntypes}(b). $E^{Pos}$ is replaced by the relative positional embedding $E^R$ that only depends on the difference between the positional indices of the array, and positional information is no longer used at key and value encoder. Parameters $b_I$ and $b_R$ serve as global content/positional bias respectively. For absolute embedding, we also adopt the previous structure. In addition, we propose separating the item and positional embedding for absolute attention as shown in Figure \ref{fig:attntypes}(c).

As explained in section \ref{section:inputembedding}, we can diversify the positional information that is fed to the attention module. MEANTIME utilizes 3 absolute embeddings and 3 relative embeddings. When fed a relative embedding, the head will operate as in Figure \ref{fig:attntypes}(b) with either $E^{Sin},E^{Exp}\text{ or }E^{Log}$ as input to $K_R$. When fed an absolute embedding, the head will operate as in Figure \ref{fig:attntypes}(c) with either $E^{Day},E^{Pos}\text{ or }E^{Con}$ as input to $Q_A\text{ and } K_A$.
As in previous works, the dimension of each head is $\frac{h}{n}$. The choice of $n$ and embedding types are adjustable with regards to the characteristics of the dataset.

\subsubsection{Stacking Layers}
After self-attention, MEANTIME operates similar to~\cite{BERT, BERT4Rec}. We apply Position-wise Feed Forward Network (FFN) to the result of self-attention to complete a single layer:
\begin{equation}
    \text{FFN}(x) = \text{GELU}(xW^1 + b^1)W^2 + b^2
\end{equation}
where $W^1 \in \mathbb{R}^{h \times 4h}, b^1 \in \mathbb{R}^{4h}, W^2 \in \mathbb{R}^{4h \times h}$ and $b^2 \in \mathbb{R}^{h}$ are learnable parameters. Then we stack $L$ such layers. As in previous works~\cite{TRANSFORMER, SASRec, BERT4Rec}, we apply a residual connection for each sublayer to facilitate training:
\begin{equation}
\begin{split}
  y &= x + \text{Dropout}(\text{Attention}(\text{LayerNorm}(x)))\\
  z &= y + \text{Dropout}(\text{FFN}(\text{LayerNorm}(y)))
\end{split}
\end{equation}
Note that while some works~\cite{TRANSFORMER, BERT4Rec} apply LayerNorm at the very end, we empirically chose to apply LayerNorm in the front. This is also in accordance with the recent report on the order of layer normalization in Transformers~\cite{LNORDER}.

\subsubsection{Prediction Layer}
Given the outputs $[{o}^1, ..., o^N] \in \mathbb{R}^{N \times h}$ from the last layer, we obtain the item score distribution for each position by:
\begin{equation}
    P\left(V | \mathbf{v}, \mathbf{t}\right) = \text{softmax}(\text{GELU}(o^{i}W^P + b^p) E^T + b^O)
\end{equation}
where $W^P \in \mathbb{R}^{h \times h}, b^P \in \mathbb{R}^{h}$, and $b^O \in \mathbb{R}^{|V|}$ are learnable parameters and $E^T \in \mathbb{R}^{h \times |V|}$ is an item embedding table. Note that we use shared item embeddings for both input and output prediction. $P\left(v_i = v | \mathbf{v}, \mathbf{t}\right)$ represents the probability that an item at position $i$ is item $v$.

\subsection{Model Training}
To train our model, we adopt the existing technique~\cite{BERT4Rec}. 
We sample a subarray of items $\mathbf{v} = [v^u_k, ..., v^u_{k+N-1}]$ and timestamps $\mathbf{t} = [t^u_k, ..., t^u_{k+N-1}]$ from $(V^u, T^u)$, and convert $\mathbf{v}$ to $\mathbf{v'}$ by randomly masking a portion of it with [MASK] by some probability $\rho$. Then we feed $\mathbf{v'}$ and $\mathbf{t}$ to our model to get
$P(V | \mathbf{v'}, \mathbf{t})$. Next, we calculate the loss as follows:
\begin{equation}
    L = -\sum_{\mathbf{v'}_i \text{ is masked}}{\log{P(v_i = \mathbf{v}_i | \mathbf{v'}, \mathbf{t})}}
\end{equation}

\section{Experiments}
\begin{table*}[t]
\small
\caption{Performance Comparison.}
\Description[The table reports the performance of all models on all four datasets]{The table reports the Recall@5, Recall@10, NDCG@5, NDCG@10 performance of all models on all datasets. It also shows the relative improvement of our model compared to the strongest baselines. Of all the models BERT is usually the strongest baseline. MEANTIME always performs better than any other baseline.}
\label{tab:results}
\begin{tabular}{@{}llrrrrrr@{}}
\toprule
Datasets & Metrics   & MARANK & SAS    & TISAS        & BERT         & MEANTIME        & Improvement \\ \midrule
ML-1M    & Recall@5  & 0.4427 & 0.5708 & \underline{0.5869} & 0.5862       & \textbf{0.6415} & 9.31\%      \\
         & Recall@10 & 0.5993 & 0.6966 & \underline{0.7102} & 0.6987       & \textbf{0.7412} & 4.36\%      \\
         & NDCG@5    & 0.3042 & 0.4142 & 0.4307       & \underline{0.4420} & \textbf{0.4932} & 11.58\%     \\
         & NDCG@10   & 0.3548 & 0.4550 & 0.4708       & \underline{0.4786} & \textbf{0.5255} & 9.81\%      \\
ML-20M   & Recall@5  & 0.3938 & 0.5274 & 0.5412       & \underline{0.5910} & \textbf{0.6225} & 5.34\%      \\
         & Recall@10 & 0.5503 & 0.6887 & 0.7021       & \underline{0.7176} & \textbf{0.7491} & 4.39\%      \\
         & NDCG@5    & 0.2665 & 0.3700 & 0.3814       & \underline{0.4469} & \textbf{0.4739} & 6.04\%      \\
         & NDCG@10   & 0.3170 & 0.4223 & 0.4336       & \underline{0.4879} & \textbf{0.5150} & 5.55\%      \\
Beauty   & Recall@5  & 0.1577 & 0.2011 & 0.2081       & \underline{0.2271} & \textbf{0.2470} & 8.78\%      \\
         & Recall@10 & 0.2302 & 0.2714 & 0.2805       & \underline{0.3095} & \textbf{0.3330} & 7.58\%      \\
         & NDCG@5    & 0.1085 & 0.1464 & 0.1512       & \underline{0.1633} & \textbf{0.1764} & 8.05\%      \\
         & NDCG@10   & 0.1318 & 0.1689 & 0.1745       & \underline{0.1898} & \textbf{0.2041} & 7.52\%      \\
Game     & Recall@5  & 0.2846 & 0.3925 & 0.3857       & \underline{0.4211} & \textbf{0.4752} & 12.85\%     \\
         & Recall@10 & 0.4217 & 0.5201 & 0.5085       & \underline{0.5590} & \textbf{0.6100} & 9.12\%      \\
         & NDCG@5    & 0.1893 & 0.2735 & 0.2735       & \underline{0.2970} & \textbf{0.3446} & 16.03\%     \\
         & NDCG@10   & 0.2334 & 0.3147 & 0.3133       & \underline{0.3415} & \textbf{0.3882} & 13.67\%     \\ \bottomrule
\end{tabular}
\end{table*}

\subsection{Setup}
\subsubsection{Datasets}
We evaluate our model on four real-word datasets with various domains and sparsity: MovieLens 1M~\cite{movielens}, MovieLens 20M~\cite{movielens}, Amazon Beauty~\cite{amazondataset} and Amazon Game~\cite{amazondataset}. We follow the common data preprocessing procedure from~\cite{TRANSREC, SASRec, TISASRec, BERT4Rec}. We convert each dataset into an implicit dataset by treating each numerical rating or review as a presence of a user-item interaction. We group the interactions by user ids and then sort them by timestamps to form a sequence for each user. Following the custom practice~\cite{TRANSREC, SASRec, TISASRec, BERT4Rec},  we discard users and items with less than five interactions to ensure the quality of the dataset.

\subsubsection{Evaluation}
To evaluate the performance of each model, we adopt the widely used \textit{leave-one-out} evaluation task. For each user's item sequence, we hold out the last item for test, and the second last item for validation. We use the rest for training. We follow the common practice~\cite{SASRec, TISASRec, BERT4Rec} of letting the model rank the ground truth item together with 100 randomly sampled negative items which haven't yet been interacted by the user. As in~\cite{BERT4Rec}, we sample the negative items according to their popularity to make the task more realistic.
To score the ranked list, we use two common metrics: \textit{Recall}, and \textit{Normalized Discounted Cumulative Gain} (NDCG). Both Recall@K and NDCG@K are designed to have larger values when the ground truth item is ranked higher up in the top-k list. We report with $K=5\text{ and }10$.

\subsection{Baselines Models \& Implementation Details}
In order to validate the effectiveness of our method, we compare it with the state-of-the-art baselines: MARank~\cite{MARANK}, SASRec~\cite{SASRec}, TiSASRec~\cite{TISASRec}, and BERT4Rec~\cite{BERT4Rec}. These models are evaluated based on the code provided by the authors.
We implemented MEANTIME\footnote{Our code is available at https://github.com/SungMinCho/MEANTIME} with \verb|PyTorch|.
All models are fully tuned through an extensive grid-search on hyperparameters: hidden dimension $h \in \{16, 32, 64, 128\}$, number of heads $n \in \{2, 4\}$, dropout ratio $\in \{0.2, 0.5\}$, and weight decay $wd \in \{0, 0.00001\}$. We used the same maximum sequence length ($N=200$ for MovieLens, $N=50$ for Amazon) and the number of layers ($L = 2$) for all transformer-based models to ensure fairness. We train all models until their validation accuracy doesn't improve for 20 epochs, and then report with the test set. We report the optimal result for each model.
As for embedding types in MEANTIME, we used the best combination of embedding types among all possible combinations with $n=2, 4$: \textbf{Day+Pos+Sin+Log} for MovieLens 1M and 20M, \textbf{Day+Sin+Exp+Log} for Amazon Beauty, and \textbf{Day+Pos+Sin+Exp} for Amazon Game.

\subsection{Results}
Table \ref{tab:results} shows the performance of all models on four datasets. It also shows the relative improvements of our model compared to the strongest baselines.
Our model gives the best performance in all metrics and datasets. On average, MEANTIME achieves \textbf{9.07\%}, \textbf{6.36\%}, \textbf{10.43\%}, and \textbf{9.14\%} improvements over the strongest baselines in Recall@5, Recall@10, NDCG@5 and NDCG@10, respectively. Given that we tune all models under the same set of hyperparameters (including the number of attention heads), we can deduce that the performance boost comes from the novel multi-temporal embeddings and attention operations that we proposed. Especially, although TiSASRec uses additional temporal information, it does not show impressive improvement over SASRec and BERT4Rec. We argue that using a single embedding scheme throughout all attention heads was a hindrance to their potential improvement. In contrast, each attention head takes different temporal embedding to extract specialized patterns in our design. This joint design of diverse temporal embeddings and attention mechanism was the key to our improvement.

\subsection{Effect of Multi-temporal Embeddings}
\begin{figure*}[t]
  \centering
  \begin{minipage}[t]{.49\linewidth}
    \centering\includegraphics[width=\linewidth]{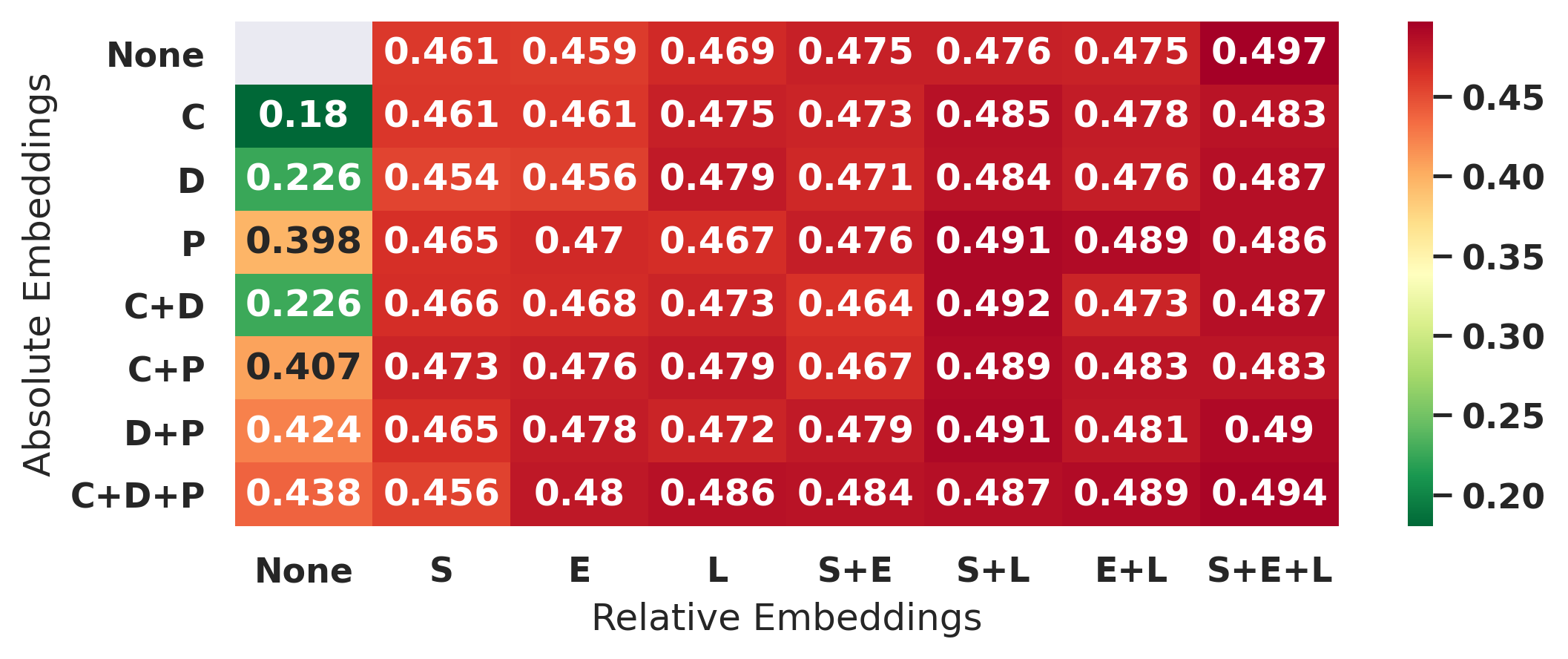}
  \end{minipage}
  \begin{minipage}[t]{.49\linewidth}
    \centering\includegraphics[width=\linewidth]{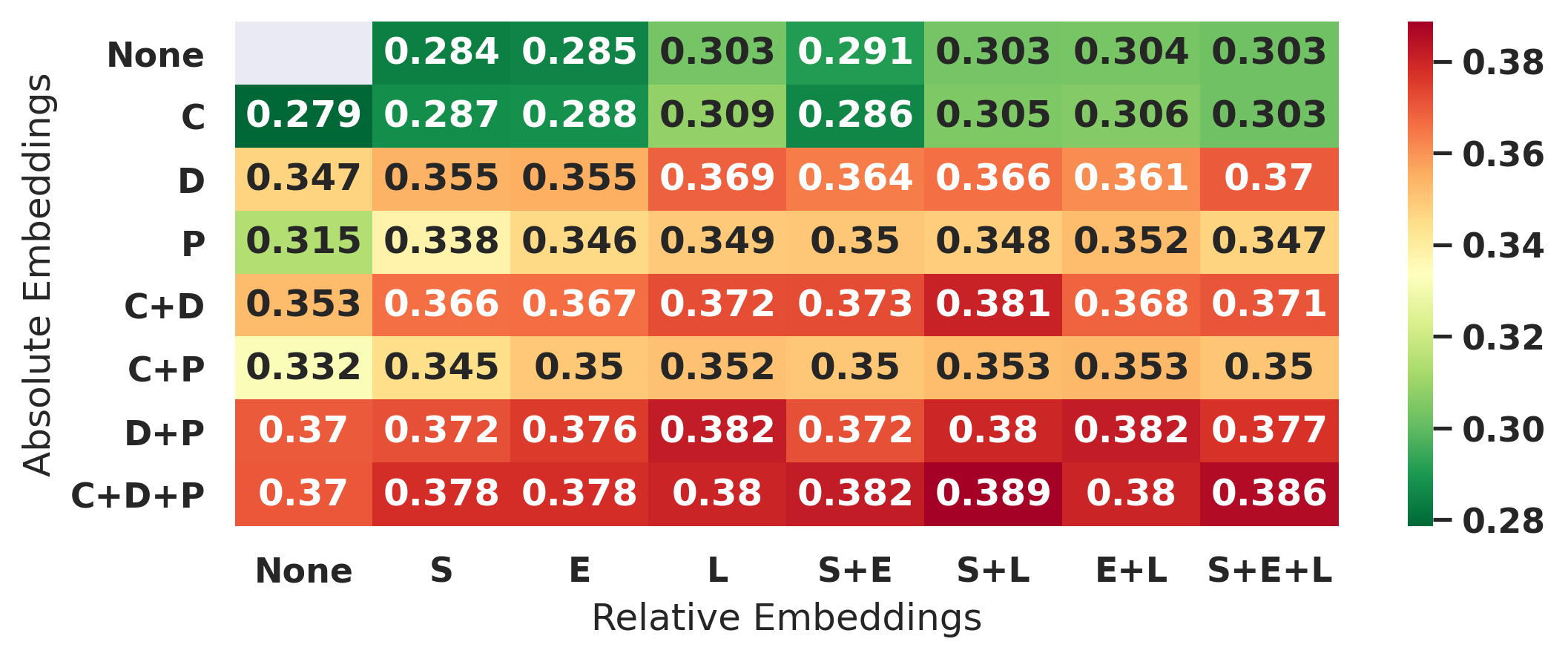}
  \end{minipage}
  \caption{NDCG@10 Comparison with All Possible Combinations of Positional Embeddings on ML-1M (left) and Game (right) Dataset. C, D, P, S, E and L represent \textbf{Con}, \textbf{Day}, \textbf{Pos}, \textbf{Sin}, \textbf{Exp} and \textbf{Log}, repectively. We used $h=60$ for all experiments.}
  \Description[The figure describes heatmaps of NDCG@10 comparison with all possible combinations of embedding types on MovieLens 1M and Amazon Game]{The left figure is a heatmap of NDCG@10 comparison with all possible combinations of embedding types on MovieLens 1M. Absolute embeddings lie on y-axis and relative embeddings lie on x-axis. As we move further into the x-axis, the heatmap gets hotter. Columns with Log embedding seems to be the hottest. The right figure describes the same heatmap on Amazon Game. As we move further down on y-axis, the heatmap gets hotter. Rows with Day embedding seems to be the hottest.}
  \label{fig:tesearch}
\end{figure*}
Figure \ref{fig:tesearch} shows the comparison of all possible combinations of temporal embeddings on ML-1M and Amazon Game dataset. We fixed all other hyperparameters (e.g., $h=60$). Interestingly, the importance of each embedding seems to differ in each dataset. In ML-1M, relative position embeddings (especially \textbf{Log}) seem to play an important role, whereas in Amazon Game, absolute position embeddings (especially \textbf{Day}) seem to be essential. Therefore, the optimal combination is also different for each dataset. Further analysis with different $h$ shows us that while optimal combination differs slightly for each $h$, the overall tendency stays the same for each dataset. The same applies to other datasets. This suggests that datasets with distinct characteristics require carefully chosen temporal embeddings that provide the right views on the context of user-item interactions. Note that while we only used $n=2\text{ and }4$ for Table \ref{tab:results} to be fair with other baselines, using all possible $n$ can further improve our performance.
\section{Conclusion}
In this paper we presented MEANTIME, which uses time information and multiple encoding functions to provide distinctive positional embeddings to self-attention modules.  Experiments on four real-world datasets show that our model outperforms state-of-the-art baselines in sequential recommendation. Through extensive ablation study, we also showed that different datasets require different positional embeddings to gain optimal performance. This suggests that one must carefully tune the positional factor in their model considering the characteristics of their own dataset.

\begin{acks}
This work was supported by Samsung Electronics and the National Research Foundation of Korea (NRF) grant (PF Class Heterogeneous High Performance Computer Development, NRF-2016M3C4A7952587) funded by the Ministry of Science, ICT \& Future Planning (No. 2013R1A3A2003664).
\end{acks}


\bibliographystyle{ACM-Reference-Format}
\bibliography{bibs}


\end{document}